\documentclass[10pt,twocolumn,letterpaper]{article}

\usepackage[applications]{wacv}
\usepackage[accsupp]{axessibility}

\definecolor{wacvblue}{rgb}{0.21,0.49,0.74}
\usepackage[pagebackref,breaklinks,colorlinks,allcolors=wacvblue]{hyperref}

\def\wacvPaperID{*****}
\def\confName{WACV}
\def\confYear{2027}

\usepackage{xcolor}

\usepackage{xspace}
\def\tool{\textsc{EDFNet}\xspace}

\usepackage{url}
\usepackage{comment}
\usepackage{amsmath}
\usepackage{enumitem}
\usepackage{graphicx}
\usepackage{hyperref}
\usepackage{multirow}
\usepackage{pgfplots}
\usepackage{adjustbox}
\usepackage{subcaption}
\pgfplotsset{compat=1.18}
\usepgfplotslibrary{colormaps}
\usepackage{tabularx,booktabs,makecell,array}
\newcolumntype{Y}{>{\raggedright\arraybackslash}X}

\title{Early-Fusion Cue Analysis for Thin-Obstacle Semantic Segmentation in UAV Navigation: A Systematic Empirical Study}

\author{
Negar Fathi\\
University of Nebraska--Lincoln\\
Lincoln, Nebraska, USA\\
{\tt\small nfathi2@huskers.unl.edu}
}

\begin{document}

\maketitle

\begin{abstract}
Thin and ultra-thin obstacles such as wires, poles, and branches remain challenging for UAV semantic segmentation because they occupy few pixels, often exhibit weak visual contrast, and are affected by severe class imbalance. Depth and edge information can complement RGB appearance with geometric and boundary cues, but their individual and combined contributions are difficult to isolate when their use is coupled with specialized fusion architectures. We present \tool, a modular early-fusion framework that enables controlled evaluation of these complementary inputs under a fixed input-level fusion strategy. We study RGB, RGB-D, RGB-E, and RGB-D-E on the Drone Depth and Obstacle Segmentation (DDOS) benchmark across four segmentation architectures under pretrained and non-pretrained settings, examining their performance through quantitative, class-wise, qualitative, and efficiency analyses. Results identify depth as the most consistently beneficial auxiliary input, whereas explicit edge information produces smaller and more heterogeneous effects. Incorporating edges alongside depth yields limited additional benefit, with the effect varying across architectures and obstacle categories. Architecture choice also has a larger influence on inference throughput than the addition of depth or edge channels.
\end{abstract}

\section{Introduction}
\label{section:introduction}
Safe autonomous navigation for unmanned aerial vehicles (UAVs) depends on reliable perception of obstacles along the flight path. Thin and ultra-thin structures such as wires, poles, branches, and fences are particularly challenging because they occupy few image pixels, often exhibit weak visual contrast against cluttered backgrounds, and may be partially occluded~\cite{madaan2017wire,zhou2017thin,kolbeinsson2024ddos}. Their small spatial footprint also leads to severe class imbalance, while spatial downsampling and feature aggregation can suppress the fine spatial evidence needed to preserve narrow structures~\cite{abdelfattah2023plgan,chen2023powerline}. These characteristics make thin-obstacle semantic segmentation an important but difficult component of UAV scene understanding.

Complementary cues can provide information that is difficult to recover from RGB appearance alone. Depth augments appearance with geometric information and has been incorporated through input-level, feature-level, gated, and representation-learning-based approaches~\cite{couprie2013indoor,hazirbas2017fusenet,chen2020sagate,yin2024dformer}, while edge and boundary information has been exploited through explicit inputs, structural branches, and boundary-oriented supervision~\cite{madaan2017wire,takikawa2019gated,li2020decoupled,li2023emnet}. However, their individual and combined contributions to UAV thin-obstacle segmentation remain insufficiently understood. Directly related aerial studies largely focus on power lines or wires~\cite{abdelfattah2020ttpla,madaan2017wire,an2023duformer,chen2023powerline,cheng2023plunext}, whereas much of the RGB-D and edge-aware literature addresses broader segmentation settings~\cite{hazirbas2017fusenet,chen2020sagate,takikawa2019gated,yin2024dformer}. Multi-cue approaches also commonly introduce dedicated fusion or cue-specific processing~\cite{zou2022ged,yang2023drnet}, making it difficult to separate cue contribution from fusion design. Consequently, a controlled characterization of depth and explicit edge information, individually and jointly, is still needed for navigation-oriented thin obstacles.

We address this gap through a controlled early-fusion study of depth and explicit edge information, considered individually and jointly relative to RGB. We present \tool, a modular framework supporting RGB, RGB-D, RGB-E, and RGB-D-E under the same input-level channel-concatenation strategy. Rather than introducing cue-specific fusion modules, the fusion mechanism is held fixed while input configuration, segmentation architecture, and initialization setting are varied. This design provides a common experimental basis for examining cue effects while controlling the fusion design and for assessing their variation across model settings and semantic categories.

We evaluate \tool on the Drone Depth and Obstacle Segmentation (DDOS) benchmark~\cite{kolbeinsson2024ddos}, whose navigation-oriented taxonomy includes \emph{Thin Structures} and \emph{Ultra-thin} obstacles together with broader scene categories. The four input configurations are evaluated with U-Net~\cite{ronneberger2015unet}, DeepLabV3~\cite{chen2017deeplabv3}, DeepLabV3+~\cite{chen2018deeplabv3plus}, and SegFormer~\cite{xie2021segformer} under pretrained and non-pretrained settings, yielding 32 model--input configurations. The evaluation examines overall, boundary, and thin-obstacle segmentation performance, class-wise cue effects, qualitative prediction behavior, and inference efficiency.

\noindent\textbf{Research Questions.} Our evaluation addresses four research questions:
\begin{list}{}{
    \setlength{\leftmargin}{2.8em}
    \setlength{\labelwidth}{2.2em}
    \setlength{\labelsep}{0.6em}
    \setlength{\itemsep}{0.2em}
    \setlength{\parsep}{0pt}
    \setlength{\topsep}{0.3em}
}
    \item[\textbf{RQ1}] \textbf{Input-cue effects:} How do depth, edge information, and their combination affect thin-obstacle segmentation relative to RGB-only input?
    
    \item[\textbf{RQ2}] \textbf{Class-wise cue effects:} How do these input cues affect different semantic classes, particularly \emph{Thin Structures} and \emph{Ultra-thin} obstacles?
    
    \item[\textbf{RQ3}] \textbf{Model and initialization effects:} How do model architecture and initialization influence segmentation performance across the evaluated input configurations?
    
    \item[\textbf{RQ4}] \textbf{Efficiency--accuracy trade-offs:} What accuracy--efficiency trade-offs arise across model and input configurations?
\end{list}

\noindent\textbf{Contributions.} The main contributions of this work are:
\begin{enumerate}[label=\textbf{\arabic*.}]
    \item We introduce a controlled early-fusion framework for UAV thin-obstacle semantic segmentation that applies a fixed input-level fusion strategy to RGB, RGB-D, RGB-E, and RGB-D-E, enabling cue effects to be examined without changes in fusion architecture.

    \item We conduct a systematic study on DDOS across four segmentation architectures and pretrained and non-pretrained settings, yielding 32 model--input configurations to assess the individual and combined effects of depth and explicit edge information relative to RGB.

    \item We provide a fine-grained characterization of cue effects across overall and boundary segmentation performance, thin-obstacle performance, paired class-wise cue effects, qualitative error patterns, and inference efficiency, establishing a controlled empirical baseline for complementary-cue evaluation.
\end{enumerate}

The remainder of the paper is organized as follows. Section~\ref{section:related-work} reviews related work, Section~\ref{section:method} presents \tool, and Section~\ref{section:experimental-setup} describes the experimental setup. Section~\ref{section:results} reports the empirical results, Section~\ref{section:discussion} discusses the findings and limitations, and Section~\ref{section:conclusion} concludes the paper and outlines future directions.

\section{Related Work}
\label{section:related-work}
Prior work relevant to UAV thin-obstacle segmentation spans aerial benchmarks, RGB-based thin-structure segmentation, RGB-D fusion, edge- and boundary-aware modeling, and multi-cue approaches combining appearance, geometric, and structural information.

\begin{figure*}[t]
    \centering
    \includegraphics[width=\textwidth]{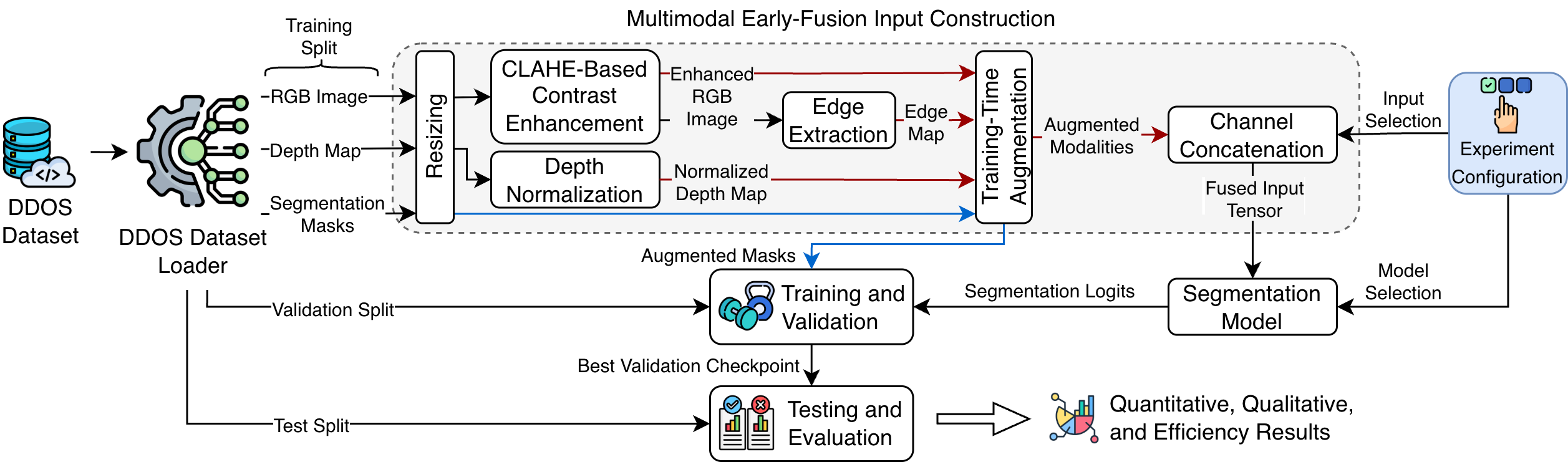}
    \caption{Overview of the \tool workflow.}
    \label{figure:workflow}
\end{figure*}

\noindent\textbf{Aerial Datasets for Thin-Structure and Obstacle Perception.}
Aerial datasets provide complementary settings for UAV scene understanding and fine-structure perception. TTPLA~\cite{abdelfattah2020ttpla} offers high-resolution imagery of transmission towers and power lines with detection, semantic segmentation, and instance segmentation annotations. UAVid~\cite{lyu2020uavid} provides densely annotated oblique urban UAV imagery with large scale variation, moving objects, and temporal consistency challenges. More directly aligned with navigation-oriented thin-obstacle perception, DDOS~\cite{kolbeinsson2024ddos} provides synthetic UAV imagery with aligned RGB, depth, semantic masks, optical flow, and surface normals, together with categories spanning ultra-thin wires and cables, thin structures such as poles and signs, and small and large mesh obstacles.

\noindent\textbf{RGB-based Thin-Structure Segmentation.}
Several RGB-based methods have been developed for aerial thin-structure segmentation. PLGAN~\cite{abdelfattah2023plgan} embeds GAN-derived decoding features into a segmentation network and introduces a Hough-space loss for very thin power lines. DUFormer~\cite{an2023duformer} combines CNN-based local feature extraction with Transformer-based global modeling on TTPLA, while PL-Deeplab~\cite{chen2023powerline} uses multibranch extraction, contextual aggregation, decoder feature fusion, and weighted loss to address complex backgrounds, boundary recovery, and class imbalance. AMFA-Net~\cite{lan2026amfanet} further combines attention, feature aggregation, and a weighted composite loss. For autonomous navigation, UCorr~\cite{kolbeinsson2024ucorr} jointly estimates wire segmentation and depth from monocular temporal imagery using correlation-based modeling. These studies remain largely centered on power lines or wires rather than broader thin-obstacle taxonomies.

\noindent\textbf{RGB-D Segmentation and Multimodal Fusion.}
RGB-D methods range from direct input-level fusion to dedicated cross-modal and representation-learning designs. Couprie et al.~\cite{couprie2013indoor} incorporate depth as an additional input channel, whereas FuseNet~\cite{hazirbas2017fusenet} uses separate RGB and depth branches and fuses depth features into the RGB stream as the encoder deepens. SA-Gate~\cite{chen2020sagate} introduces separation-and-aggregation gating with bidirectional cross-modality propagation, and DFormer~\cite{yin2024dformer} learns transferable RGB-D representations through pretraining. In aerial imagery, EDFT~\cite{yan2022edft} uses depth-aware self-attention, while Chao et al.~\cite{chao2026m3wavegnet} introduce the simulated UAV RGB-D APSD dataset and M$^3$WaveGNet, a real-time model using multi-level wavelets and multimodal fusion. Across this literature, depth effects are commonly evaluated within different fusion or representation designs.

\noindent\textbf{Edge- and Boundary-aware Segmentation.}
Edge and boundary information has been incorporated through structural streams, auxiliary supervision, and explicit input cues. Gated-SCNN~\cite{takikawa2019gated} introduces a gated shape stream and reports particularly strong gains for thinner and smaller objects. Zhou et al.~\cite{zhou2017thin} represent wires, cables, and branches using image edges and reconstruct their 3D locations from monocular or stereo video. Madaan et al.~\cite{madaan2017wire} formulate UAV wire detection as semantic segmentation and evaluate edge- and line-detector outputs as additional input channels. In aerial segmentation, EMNet~\cite{li2023emnet} adds an encoder edge-detection branch with multi-level upsampling, while PL-UNeXt~\cite{cheng2023plunext} uses edge-detail and line-feature heads to guide lower- and higher-level learning. Thus, structural information is used in several forms, but in many methods its effect is coupled with auxiliary branches, losses, or architecture-specific processing.

\noindent\textbf{Appearance--Geometry--Edge and Multi-Cue Fusion.}
Multi-cue approaches further combine appearance, geometry, and edge-related information through specialized designs. Gate-Guided Edge Distillation~\cite{zou2022ged} adaptively fuses RGB and depth to generate edge-aware features that guide semantic prediction. FCEGNet~\cite{lu2025fcegnet} combines cross-modal feature calibration with an edge-guided decoder that enhances edge information while preserving spatial information. In remote sensing, BATFNet~\cite{tong2026batfnet} integrates RGB and DSM with DSM-derived edge priors for bidirectional cross-modal attention and decoder refinement. These methods motivate joint use of geometric and structural cues, but evaluate them together with method-specific fusion mechanisms.

Taken together, prior work establishes strong foundations for aerial thin-structure segmentation, RGB-D fusion, and edge-aware modeling, but does not directly provide the controlled comparison pursued here. Directly related aerial studies largely emphasize power lines or wires, while RGB-D and edge-aware methods often target broader segmentation settings or employ specialized cue-processing mechanisms. Our study instead uses DDOS and holds the input-level fusion mechanism fixed across cue configurations, architectures, and initialization settings, enabling depth and explicit edge information to be compared under a common experimental design.

\section{Method}
\label{section:method}
This section presents the \tool framework, including its workflow, early-fusion input construction, and segmentation architectures.

\subsection{Overview of \textbf{\tool}}
\label{subsection:overview}
Figure~\ref{figure:workflow} summarizes the \tool workflow. The DDOS loader provides aligned RGB images, depth maps, and segmentation masks. During multimodal input construction, the inputs are resized, RGB is enhanced using CLAHE~\cite{zuiderveld1994clahe}, depth is normalized, and edges are extracted from the enhanced RGB image. Training-time augmentation is then applied before the experiment configuration selects the input combination for channel-wise concatenation into a fused input tensor. The selected segmentation model is trained and validated, and the retained validation checkpoint is evaluated on the held-out test split to produce quantitative, qualitative, and efficiency results. This common pipeline enables consistent comparison across input configurations, model architectures, and initialization settings.

\subsection{Early-Fusion Input Configurations}
\label{subsection:input-configurations}
\tool evaluates four input configurations: RGB, RGB-D, RGB-E, and RGB-D-E. RGB provides the three-channel appearance input, while depth and edge maps provide geometric and structural cues, respectively. RGB-D and RGB-E therefore contain four channels, whereas RGB-D-E combines all three cues in a five-channel representation. In every case, the selected channels are concatenated before feature extraction, keeping the early-fusion operation identical across configurations.

Aligned RGB images, depth maps, and segmentation masks are first resized to $256\times256$ pixels, with nearest-neighbor interpolation used for segmentation masks. RGB images are then enhanced channel-wise using CLAHE, and edge maps are obtained from the enhanced RGB by grayscale conversion followed by Sobel gradient extraction~\cite{sobel1968}. The resulting edge magnitudes are normalized to $[0,1]$. DDOS depth maps are converted to floating point, scaled using the fixed divisor $65535$, and clipped to $[0,1]$, preserving a common depth scale across samples. During training, spatial augmentations---horizontal flipping, shift--scale--rotation, and random resized cropping---are applied jointly to RGB, depth, edge, and segmentation masks to maintain pixel-level alignment. Photometric augmentation is applied only to RGB using random brightness/contrast adjustment and stochastic CLAHE; because these operations alter image appearance, the edge map is recomputed from the augmented RGB image before fusion. Validation and test samples are processed without augmentation.

The resulting channels are concatenated to form the model input. For non-pretrained models, the constructed inputs retain their $[0,1]$ scaling. For pretrained models, the RGB channels are standardized using ImageNet statistics~\cite{deng2009imagenet}, while additional depth and edge channels are mapped from $[0,1]$ to $[-1,1]$. This preserves the same input-level fusion design across all configurations while adapting input normalization to the initialization setting.

\subsection{Segmentation Architectures}
\label{subsection:segmentation-architectures}

To examine whether cue effectiveness depends on model architecture, \tool evaluates four segmentation families: U-Net~\cite{ronneberger2015unet}, DeepLabV3~\cite{chen2017deeplabv3}, DeepLabV3+~\cite{chen2018deeplabv3plus}, and SegFormer~\cite{xie2021segformer}. These models span encoder--decoder, atrous-convolutional, and transformer-based designs, providing different mechanisms for preserving spatial detail and modeling context. Both pretrained and non-pretrained settings are evaluated for each family.

The non-pretrained U-Net is a custom four-level encoder--decoder with skip connections, whereas the pretrained U-Net uses a ResNet-34 encoder~\cite{he2016resnet} initialized from ImageNet; consequently, the two U-Net settings are not architecture-matched. DeepLabV3 uses a ResNet-50 backbone~\cite{he2016resnet} with either random initialization or pretrained DeepLabV3-ResNet50 weights. DeepLabV3+ uses a ResNet-34 encoder with either random or ImageNet-pretrained initialization, while SegFormer uses the hierarchical MiT-B0 encoder with either non-pretrained or ImageNet-pretrained weights.

Each model is configured to accept the channel dimensionality of the selected RGB, RGB-D, RGB-E, or RGB-D-E input. For DeepLabV3, the first ResNet convolution is explicitly adapted for additional channels; in the pretrained setting, the original RGB kernels are retained, while each added channel is initialized from the mean of the pretrained RGB kernels. U-Net, DeepLabV3+, and SegFormer receive the required input-channel count directly through their model constructors. This allows all four input configurations to be evaluated without introducing cue-specific fusion modules, preserving the same input-level early-fusion design across architectures.

\section{Experimental Setup}
\label{section:experimental-setup}
This section describes the dataset, training protocol, and evaluation measures used to evaluate \tool.

\subsection{Dataset}
\label{subsection:dataset}
Experiments are conducted on the Drone Depth and Obstacle Segmentation (DDOS) dataset~\cite{kolbeinsson2024ddos}, a synthetic UAV benchmark containing 34{,}000 frames from 340 simulated flights at $1280\times720$ resolution. We use the official training, validation, and test partitions of 300, 20, and 20 flights, respectively. DDOS defines ten semantic classes: Background, Other, Animals, Vehicles, Buildings, Trees, Large Mesh, Small Mesh, Thin Structures, and Ultra-thin. Our thin-obstacle analysis focuses on the target set $\mathcal{T}=\{\text{Thin Structures},\text{Ultra-thin}\}$. Only complete frame-aligned RGB, depth, and segmentation-mask triplets are retained. Input construction and preprocessing follow Section~\ref{subsection:input-configurations}.

\subsection{Training Protocol}
\label{subsection:training-protocol}
We evaluate RGB, RGB-D, RGB-E, and RGB-D-E with U-Net, DeepLabV3, DeepLabV3+, and SegFormer under pretrained and non-pretrained settings, yielding 32 model--input configurations. Each configuration is trained independently for 50 epochs using Adam~\cite{kingma2015adam}, a learning rate of $5\times10^{-4}$, and a batch size of 16, with no learning-rate scheduler; the same batch size is used for validation and testing. All trainable parameters are optimized end-to-end using class-weighted cross-entropy. For class $c$ with $n_c$ training pixels, the initial weight is $w_c=N/(C n_c)$, where $N$ is the total number of labeled training pixels and $C$ is the number of classes. Weights for classes present in the training split are clipped at 10, normalized by their mean, and kept fixed during optimization; absent classes receive zero weight.

The training-time augmentations described in Section~\ref{subsection:input-configurations} include horizontal flipping with probability $0.5$; shift--scale--rotation with limits $0.05$, $0.1$, and $10^\circ$, respectively, with probability $0.5$; and random resized cropping with scale $[0.8,1.0]$ and aspect ratio $[0.9,1.1]$, also with probability $0.5$. RGB-only photometric augmentation applies random brightness/contrast adjustment with probability $0.5$ and CLAHE with probability $0.3$.

Validation is performed after every epoch. Checkpoints are ranked using a fixed validation rule that emphasizes thin-obstacle performance by combining mean IoU over \emph{Thin Structures} and \emph{Ultra-thin}, mean thin-class $F_2$, mBIoU, mIoU, and mean thin-class precision (weights $0.35$, $0.30$, $0.20$, $0.10$, and $0.05$, respectively). The same rule is applied to all configurations, and the highest-scoring validation checkpoint is retained for final evaluation on the held-out test split. Experiments use random seed 42 with deterministic Python, NumPy, PyTorch, and cuDNN settings where supported, and are conducted on an NVIDIA RTX 5000 Ada Generation GPU with 32~GB of memory.

\subsection{Evaluation Metrics}
\label{subsection:evaluation-metrics}
Segmentation performance is computed over the complete test split using a dataset-level confusion matrix. We report mean Intersection-over-Union (mIoU)~\cite{everingham2010pascal}, defined as the macro average of class-wise IoU over valid semantic classes, together with $\mathrm{IoU}_{\mathrm{Thin}}$ and $\mathrm{IoU}_{\mathrm{Ultra}}$ for the two target thin-obstacle categories.

For the target set $\mathcal{T}$, class-wise precision, recall, and $F_2$ are macro-averaged to obtain $\overline{P}_{\mathcal{T}}$, $\overline{R}_{\mathcal{T}}$, and $\overline{F}_{2,\mathcal{T}}$. For class $c$, the recall-oriented $F_2$ score is
\begin{equation}
F_{2,c}
=
\frac{5P_cR_c}{4P_c+R_c},
\label{equation:f2}
\end{equation}
where $P_c$ and $R_c$ denote precision and recall. Compared with $F_1$, $F_2$ places greater emphasis on recall, reflecting the importance of missed thin-obstacle pixels. Classes absent from the evaluated data are excluded from the corresponding macro average; if a present thin class receives no predicted pixels, its $F_2$ contribution is set to zero.

Boundary quality is evaluated using Boundary IoU~\cite{cheng2021boundaryiou}. Predicted and ground-truth boundary regions are constructed per class using a $3\times3$ morphological kernel with boundary width set to $2\%$ of the image diagonal, and mBIoU denotes the macro average of class-wise Boundary IoU over valid classes.

For the class-wise cue analysis in Fig.~\ref{figure:class-wise-results}, $\Delta\mathrm{IoU}$ is computed between paired input configurations while holding model architecture and initialization fixed. Each value is the IoU of the first configuration minus that of the second, so positive values favor the first configuration.

Computational efficiency is reported as inference throughput in frames per second (FPS). CUDA execution is synchronized immediately before and after each model forward pass, and FPS is computed from batch size and mean forward-pass time. The reported throughput therefore measures model inference under a common hardware and evaluation protocol, excluding data loading and input preprocessing.

\section{Results}
\label{section:results}
This section presents quantitative, qualitative, and efficiency analyses of the evaluated model--input configurations on the DDOS test split.

\begin{table*}[t]
\centering
\caption{Aggregate quantitative results by input configuration and model setting on the DDOS test split. Subtable~(a) averages over model settings, while Subtable~(b) averages over input configurations. Higher values indicate better performance. Bold denotes the best value in each subtable; \emph{Pre} denotes pretrained initialization.}
\label{table:aggregate-results}

\scriptsize
\setlength{\tabcolsep}{2.8pt}
\renewcommand{\arraystretch}{1.05}

\begin{subtable}[t]{0.495\textwidth}
\centering
\caption{Input-configuration averages over model settings.}
\label{table:aggregate-input}

\resizebox{\linewidth}{!}{%
\begin{tabular}{l cccccc}
\toprule
\textbf{Input} &
\textbf{mIoU} &
\textbf{mBIoU} &
$\mathbf{IoU}_{\mathrm{Thin}}$ &
$\mathbf{IoU}_{\mathrm{Ultra}}$ &
$\mathbf{\overline{F}_{2,\mathcal{T}}}$ &
\textbf{FPS} \\
\midrule

RGB
& 0.4314
& 0.5158
& 0.2068
& 0.0789
& 0.3668
& 647.4 \\

RGB-D
& 0.4570
& 0.5781
& \textbf{0.2388}
& 0.1185
& \textbf{0.4602}
& 688.6 \\

RGB-E
& 0.4317
& 0.5171
& 0.2072
& 0.0782
& 0.3671
& \textbf{690.6} \\

RGB-D-E
& \textbf{0.4604}
& \textbf{0.5807}
& 0.2343
& \textbf{0.1203}
& 0.4593
& 688.4 \\

\bottomrule
\end{tabular}%
}
\end{subtable}
\hfill
\begin{subtable}[t]{0.495\textwidth}
\centering
\caption{Model-setting averages over input configurations.}
\label{table:aggregate-model}

\resizebox{\linewidth}{!}{%
\begin{tabular}{l cccccc}
\toprule
\textbf{Model Setting} &
\textbf{mIoU} &
\textbf{mBIoU} &
$\mathbf{IoU}_{\mathrm{Thin}}$ &
$\mathbf{IoU}_{\mathrm{Ultra}}$ &
$\mathbf{\overline{F}_{2,\mathcal{T}}}$ &
\textbf{FPS} \\
\midrule

U-Net
& \textbf{0.5104}
& \textbf{0.6091}
& \textbf{0.3197}
& \textbf{0.1452}
& \textbf{0.5243}
& 129.6 \\

U-Net (Pre)
& 0.4578
& 0.5543
& 0.2593
& 0.1074
& 0.4493
& \textbf{1419.5} \\

DeepLabV3
& 0.4286
& 0.5445
& 0.1869
& 0.0861
& 0.3786
& 371.3 \\

DeepLabV3 (Pre)
& 0.4352
& 0.5619
& 0.1924
& 0.0896
& 0.3907
& 343.0 \\

DeepLabV3+
& 0.4286
& 0.5210
& 0.2029
& 0.0877
& 0.3874
& 1081.3 \\

DeepLabV3+ (Pre)
& 0.4412
& 0.5456
& 0.2175
& 0.0936
& 0.4071
& 1038.7 \\

SegFormer
& 0.4245
& 0.5125
& 0.1920
& 0.0875
& 0.3742
& 522.0 \\

SegFormer (Pre)
& 0.4346
& 0.5346
& 0.2037
& 0.0945
& 0.3951
& 524.6 \\

\bottomrule
\end{tabular}%
}
\end{subtable}

\end{table*}

\begin{table*}[t]
\centering
\caption{Quantitative results for all 32 model--input configurations on the DDOS test split. Higher values indicate better performance. Bold denotes the best value in each column; \emph{Pre} denotes pretrained initialization.}
\label{table:full-results}

\scriptsize
\setlength{\tabcolsep}{3.2pt}
\renewcommand{\arraystretch}{1.05}

\resizebox{0.91\textwidth}{!}{%
\begin{tabular}{ll cc ccccc c}
\toprule
\multirow{2}{*}{\textbf{Model}} &
\multirow{2}{*}{\textbf{Input}} &
\multicolumn{2}{c}{\textbf{Overall Segmentation}} &
\multicolumn{5}{c}{\textbf{Thin-Obstacle Performance}} &
\multicolumn{1}{c}{\textbf{Efficiency}} \\
\cmidrule(lr){3-4}
\cmidrule(lr){5-9}
\cmidrule(lr){10-10}
&
&
\textbf{mIoU} &
\textbf{mBIoU} &
$\mathbf{IoU}_{\mathrm{Thin}}$ &
$\mathbf{IoU}_{\mathrm{Ultra}}$ &
$\mathbf{\overline{P}_{\mathcal{T}}}$ &
$\mathbf{\overline{R}_{\mathcal{T}}}$ &
$\mathbf{\overline{F}_{2,\mathcal{T}}}$ &
\textbf{FPS} \\
\midrule

\multirow{4}{*}{U-Net}
& RGB     & 0.4920 & 0.5754 & 0.3021 & 0.1148 & 0.2278 & 0.6538 & 0.4705 & 129.7 \\
& RGB-D   & \textbf{0.5295} & 0.6419 & \textbf{0.3523} & 0.1739 &
\textbf{0.2749} & \textbf{0.8555} & \textbf{0.5866} & 129.5 \\
& RGB-E   & 0.4912 & 0.5758 & 0.2959 & 0.1116 & 0.2242 & 0.6332 & 0.4595 & 129.7 \\
& RGB-D-E & 0.5290 & \textbf{0.6432} & 0.3284 & \textbf{0.1806} &
0.2655 & 0.8474 & 0.5807 & 129.4 \\
\addlinespace

\multirow{4}{*}{U-Net (Pre)}
& RGB     & 0.4423 & 0.5242 & 0.2381 & 0.0876 & 0.1784 & 0.5891 & 0.4012 & \textbf{1426.8} \\
& RGB-D   & 0.4690 & 0.5811 & 0.2807 & 0.1247 & 0.2139 & 0.7881 & 0.4978 & 1420.9 \\
& RGB-E   & 0.4429 & 0.5308 & 0.2490 & 0.0731 & 0.1759 & 0.5935 & 0.3937 & 1422.2 \\
& RGB-D-E & 0.4769 & 0.5810 & 0.2694 & 0.1441 & 0.2194 & 0.7625 & 0.5044 & 1408.4 \\
\addlinespace

\multirow{4}{*}{DeepLabV3}
& RGB     & 0.4180 & 0.5165 & 0.1803 & 0.0671 & 0.1343 & 0.5590 & 0.3401 & 372.0 \\
& RGB-D   & 0.4371 & 0.5727 & 0.1946 & 0.1041 & 0.1574 & 0.7362 & 0.4182 & 371.5 \\
& RGB-E   & 0.4191 & 0.5115 & 0.1827 & 0.0744 & 0.1414 & 0.5386 & 0.3444 & 371.6 \\
& RGB-D-E & 0.4401 & 0.5773 & 0.1899 & 0.0989 & 0.1514 & 0.7497 & 0.4117 & 369.8 \\
\addlinespace

\multirow{4}{*}{DeepLabV3 (Pre)}
& RGB     & 0.4301 & 0.5383 & 0.1872 & 0.0759 & 0.1433 & 0.5782 & 0.3562 & 341.7 \\
& RGB-D   & 0.4408 & 0.5832 & 0.1954 & 0.1053 & 0.1579 & 0.7527 & 0.4226 & 343.1 \\
& RGB-E   & 0.4274 & 0.5401 & 0.1854 & 0.0766 & 0.1415 & 0.6008 & 0.3604 & 343.7 \\
& RGB-D-E & 0.4426 & 0.5861 & 0.2015 & 0.1005 & 0.1582 & 0.7686 & 0.4238 & 343.4 \\
\addlinespace

\multirow{4}{*}{DeepLabV3+}
& RGB     & 0.4080 & 0.4797 & 0.1871 & 0.0682 & 0.1404 & 0.5356 & 0.3403 & 1044.8 \\
& RGB-D   & 0.4403 & 0.5539 & 0.2206 & 0.1071 & 0.1745 & 0.7238 & 0.4346 & 1079.1 \\
& RGB-E   & 0.4171 & 0.4903 & 0.1834 & 0.0677 & 0.1373 & 0.5403 & 0.3389 & 1112.5 \\
& RGB-D-E & 0.4490 & 0.5599 & 0.2205 & 0.1078 & 0.1743 & 0.7204 & 0.4358 & 1088.6 \\
\addlinespace

\multirow{4}{*}{DeepLabV3+ (Pre)}
& RGB     & 0.4292 & 0.5165 & 0.2040 & 0.0706 & 0.1501 & 0.5728 & 0.3602 & 826.1 \\
& RGB-D   & 0.4511 & 0.5710 & 0.2272 & 0.1139 & 0.1810 & 0.7310 & 0.4477 & 1117.6 \\
& RGB-E   & 0.4314 & 0.5204 & 0.2070 & 0.0765 & 0.1551 & 0.5747 & 0.3689 & 1099.8 \\
& RGB-D-E & 0.4532 & 0.5744 & 0.2317 & 0.1133 & 0.1828 & 0.7408 & 0.4515 & 1111.1 \\
\addlinespace

\multirow{4}{*}{SegFormer}
& RGB     & 0.4077 & 0.4733 & 0.1763 & 0.0705 & 0.1395 & 0.4766 & 0.3212 & 517.7 \\
& RGB-D   & 0.4439 & 0.5556 & 0.2092 & 0.1077 & 0.1681 & 0.7278 & 0.4292 & 522.0 \\
& RGB-E   & 0.4067 & 0.4671 & 0.1688 & 0.0680 & 0.1313 & 0.5039 & 0.3209 & 520.1 \\
& RGB-D-E & 0.4396 & 0.5539 & 0.2136 & 0.1038 & 0.1690 & 0.7140 & 0.4254 & 528.2 \\
\addlinespace

\multirow{4}{*}{SegFormer (Pre)}
& RGB     & 0.4237 & 0.5024 & 0.1791 & 0.0763 & 0.1403 & 0.5415 & 0.3443 & 520.2 \\
& RGB-D   & 0.4445 & 0.5657 & 0.2305 & 0.1109 & 0.1820 & 0.7307 & 0.4452 & 525.0 \\
& RGB-E   & 0.4176 & 0.5009 & 0.1856 & 0.0776 & 0.1453 & 0.5483 & 0.3498 & 525.2 \\
& RGB-D-E & 0.4526 & 0.5695 & 0.2195 & 0.1133 & 0.1768 & 0.7247 & 0.4410 & 527.8 \\

\bottomrule
\end{tabular}%
}
\end{table*}

\begin{figure*}[t]
\centering

\pgfplotsset{
cueheatmap/.style={
    width=4.00cm,
    height=3.35cm,
    scale only axis,
    axis on top,
    enlargelimits=false,
    xmin=-0.5,
    xmax=9.5,
    ymin=-0.5,
    ymax=7.5,
    xtick={0,1,2,3,4,5,6,7,8,9},
    xticklabels={
        {Background},
        {Other},
        {Animals},
        {Vehicles},
        {Buildings},
        {Trees},
        {Large Mesh},
        {Small Mesh},
        {Thin Structures},
        {Ultra-thin}
    },
    ytick={0,1,2,3,4,5,6,7},
    y dir=reverse,
    tick style={draw=none},
    axis line style={draw=none},
    x tick label style={font=\scriptsize,rotate=67,anchor=east},
    y tick label style={font=\scriptsize},
    colormap={cueeffect}{
        color(0cm)=(blue!75!black);
        color(1cm)=(white);
        color(2cm)=(red!75!black)
    },
    point meta min=-0.10,
    point meta max=0.10
}
}

\noindent\makebox[\textwidth][c]{%

\begin{minipage}[t]{0.31\textwidth}
\centering
\scalebox{0.88}{%
\begin{tikzpicture}
\begin{axis}[
    cueheatmap,
    yticklabels={
        {U-Net},
        {U-Net (Pre)},
        {DeepLabV3},
        {DeepLabV3 (Pre)},
        {DeepLabV3+},
        {DeepLabV3+ (Pre)},
        {SegFormer},
        {SegFormer (Pre)}
    }
]

\addplot[
    matrix plot*,
    mesh/cols=10,
    point meta=explicit
]
table[row sep=\\,meta=value] {
x y value\\
0 0 0.012654\\
1 0 0.052351\\
2 0 0.038777\\
3 0 0.041041\\
4 0 0.023834\\
5 0 0.010394\\
6 0 0.034473\\
7 0 0.051600\\
8 0 0.050129\\
9 0 0.059061\\
0 1 0.011048\\
1 1 0.046384\\
2 1 0.013532\\
3 1 0.020570\\
4 1 0.020783\\
5 1 0.005798\\
6 1 0.034760\\
7 1 0.034702\\
8 1 0.042580\\
9 1 0.037150\\
0 2 0.009921\\
1 2 0.006164\\
2 2 0.009443\\
3 2 0.057463\\
4 2 0.021940\\
5 2 0.005624\\
6 2 0.015582\\
7 2 0.013229\\
8 2 0.014306\\
9 2 0.036960\\
0 3 0.007524\\
1 3 -0.007053\\
2 3 -0.005272\\
3 3 0.010205\\
4 3 0.023468\\
5 3 0.005766\\
6 3 -0.005641\\
7 3 0.040287\\
8 3 0.008181\\
9 3 0.029388\\
0 4 0.019389\\
1 4 0.032169\\
2 4 0.004780\\
3 4 0.093587\\
4 4 0.032070\\
5 4 0.012694\\
6 4 0.015963\\
7 4 0.040333\\
8 4 0.033486\\
9 4 0.038904\\
0 5 0.013869\\
1 5 0.004018\\
2 5 0.001547\\
3 5 0.030888\\
4 5 0.026979\\
5 5 0.009058\\
6 5 0.026183\\
7 5 0.039587\\
8 5 0.023247\\
9 5 0.043304\\
0 6 0.019075\\
1 6 0.059121\\
2 6 0.000590\\
3 6 0.092975\\
4 6 0.038243\\
5 6 0.010244\\
6 6 0.011660\\
7 6 0.060520\\
8 6 0.032952\\
9 6 0.037201\\
0 7 0.014590\\
1 7 0.017611\\
2 7 -0.006461\\
3 7 0.023832\\
4 7 0.024987\\
5 7 0.009189\\
6 7 0.001660\\
7 7 0.036659\\
8 7 0.051427\\
9 7 0.034628\\
};

\draw[black,thick]
(axis cs:7.5,-0.5) -- (axis cs:7.5,7.5);

\end{axis}
\end{tikzpicture}%
}

\par\vspace{1mm}
{\footnotesize (a) RGB-D $-$ RGB}
\end{minipage}%
\hfill

\begin{minipage}[t]{0.31\textwidth}
\centering
\scalebox{0.88}{%
\begin{tikzpicture}
\begin{axis}[
    cueheatmap,
    yticklabels={}
]

\addplot[
    matrix plot*,
    mesh/cols=10,
    point meta=explicit
]
table[row sep=\\,meta=value] {
x y value\\
0 0 0.000526\\
1 0 -0.001630\\
2 0 0.019451\\
3 0 0.004853\\
4 0 -0.005609\\
5 0 -0.002859\\
6 0 -0.021063\\
7 0 0.007402\\
8 0 -0.006253\\
9 0 -0.003254\\
0 1 0.001481\\
1 1 -0.006163\\
2 1 -0.003318\\
3 1 -0.004288\\
4 1 -0.001909\\
5 1 -0.006869\\
6 1 0.018268\\
7 1 0.012286\\
8 1 0.010864\\
9 1 -0.014455\\
0 2 0.000942\\
1 2 0.005351\\
2 2 0.000480\\
3 2 -0.001325\\
4 2 -0.003924\\
5 2 -0.001016\\
6 2 0.012974\\
7 2 -0.011900\\
8 2 0.002400\\
9 2 0.007312\\
0 3 -0.001418\\
1 3 -0.013438\\
2 3 -0.009320\\
3 3 -0.019563\\
4 3 0.004883\\
5 3 0.000423\\
6 3 -0.010502\\
7 3 0.023246\\
8 3 -0.001852\\
9 3 0.000725\\
0 4 0.005580\\
1 4 0.009905\\
2 4 -0.006122\\
3 4 0.059092\\
4 4 0.000765\\
5 4 0.000746\\
6 4 0.012197\\
7 4 0.013621\\
8 4 -0.003721\\
9 4 -0.000583\\
0 5 0.001742\\
1 5 0.008723\\
2 5 -0.002303\\
3 5 0.006289\\
4 5 -0.001285\\
5 5 0.000594\\
6 5 -0.006122\\
7 5 0.005432\\
8 5 0.002998\\
9 5 0.005942\\
0 6 -0.003054\\
1 6 0.001677\\
2 6 -0.006767\\
3 6 0.013055\\
4 6 0.004450\\
5 6 0.000996\\
6 6 -0.007029\\
7 6 -0.003171\\
8 6 -0.007438\\
9 6 -0.002537\\
0 7 -0.000657\\
1 7 -0.002508\\
2 7 -0.015749\\
3 7 -0.013434\\
4 7 -0.002251\\
5 7 -0.001835\\
6 7 -0.021595\\
7 7 -0.010888\\
8 7 0.006505\\
9 7 0.001304\\
};

\draw[black,thick]
(axis cs:7.5,-0.5) -- (axis cs:7.5,7.5);

\end{axis}
\end{tikzpicture}%
}

\par\vspace{1mm}
{\footnotesize (b) RGB-E $-$ RGB}
\end{minipage}%
\hfill

\begin{minipage}[t]{0.31\textwidth}
\centering
\scalebox{0.88}{%
\begin{tikzpicture}
\begin{axis}[
    cueheatmap,
    yticklabels={},
    colorbar,
    colorbar style={
        title={$\Delta\mathrm{IoU}$},
        title style={font=\scriptsize},
        yticklabel style={font=\scriptsize},
        width=0.15cm,
        height=2.65cm,
        xshift=-1.5mm
    }
]

\addplot[
    matrix plot*,
    mesh/cols=10,
    point meta=explicit
]
table[row sep=\\,meta=value] {
x y value\\
0 0 -0.000777\\
1 0 -0.016920\\
2 0 0.021144\\
3 0 0.010396\\
4 0 0.000451\\
5 0 -0.000957\\
6 0 -0.002777\\
7 0 0.001710\\
8 0 -0.023920\\
9 0 0.006647\\
0 1 0.000368\\
1 1 0.014994\\
2 1 -0.010539\\
3 1 0.031795\\
4 1 -0.000032\\
5 1 0.002824\\
6 1 0.038294\\
7 1 -0.006452\\
8 1 -0.011351\\
9 1 0.019402\\
0 2 0.000677\\
1 2 0.030068\\
2 2 -0.004323\\
3 2 -0.005689\\
4 2 -0.000199\\
5 2 0.000842\\
6 2 0.009794\\
7 2 0.008640\\
8 2 -0.004791\\
9 2 -0.005144\\
0 3 0.000843\\
1 3 0.016962\\
2 3 -0.002153\\
3 3 0.001143\\
4 3 0.000476\\
5 3 0.001685\\
6 3 -0.001142\\
7 3 -0.001198\\
8 3 0.006107\\
9 3 -0.004814\\
0 4 0.005284\\
1 4 0.012541\\
2 4 0.008860\\
3 4 0.016520\\
4 4 0.003414\\
5 4 0.000378\\
6 4 0.020437\\
7 4 0.018942\\
8 4 -0.000136\\
9 4 0.000660\\
0 5 0.000394\\
1 5 0.004644\\
2 5 -0.003226\\
3 5 -0.013577\\
4 5 0.000486\\
5 5 0.001112\\
6 5 0.009984\\
7 5 0.017036\\
8 5 0.004477\\
9 5 -0.000519\\
0 6 -0.000186\\
1 6 -0.019769\\
2 6 0.009640\\
3 6 -0.021350\\
4 6 -0.000476\\
5 6 0.002383\\
6 6 0.000917\\
7 6 -0.014346\\
8 6 0.004366\\
9 6 -0.003977\\
0 7 0.000528\\
1 7 0.024573\\
2 7 0.007121\\
3 7 0.044579\\
4 7 0.004509\\
5 7 0.003064\\
6 7 0.001993\\
7 7 0.002600\\
8 7 -0.011082\\
9 7 0.002401\\
};

\draw[black,thick]
(axis cs:7.5,-0.5) -- (axis cs:7.5,7.5);

\end{axis}
\end{tikzpicture}%
}

\par\vspace{1mm}
{\footnotesize (c) RGB-D-E $-$ RGB-D}
\end{minipage}%
}

\caption{
Class-wise $\Delta\mathrm{IoU}$ between paired input configurations on the DDOS test split.
Within each panel, $\Delta\mathrm{IoU}$ is the IoU of the first configuration minus that of the second, with segmentation architecture and initialization fixed; positive values therefore favor the first configuration.
All panels use the same symmetric scale $[-0.10,0.10]$; the vertical separator marks \emph{Thin Structures} and \emph{Ultra-thin}.
\emph{Pre} denotes pretrained initialization.
}
\label{figure:class-wise-results}

\end{figure*}

\begin{figure*}[t]
    \centering


    \begin{subfigure}[t]{0.19\textwidth}
        \centering
        \includegraphics[width=\linewidth]
        {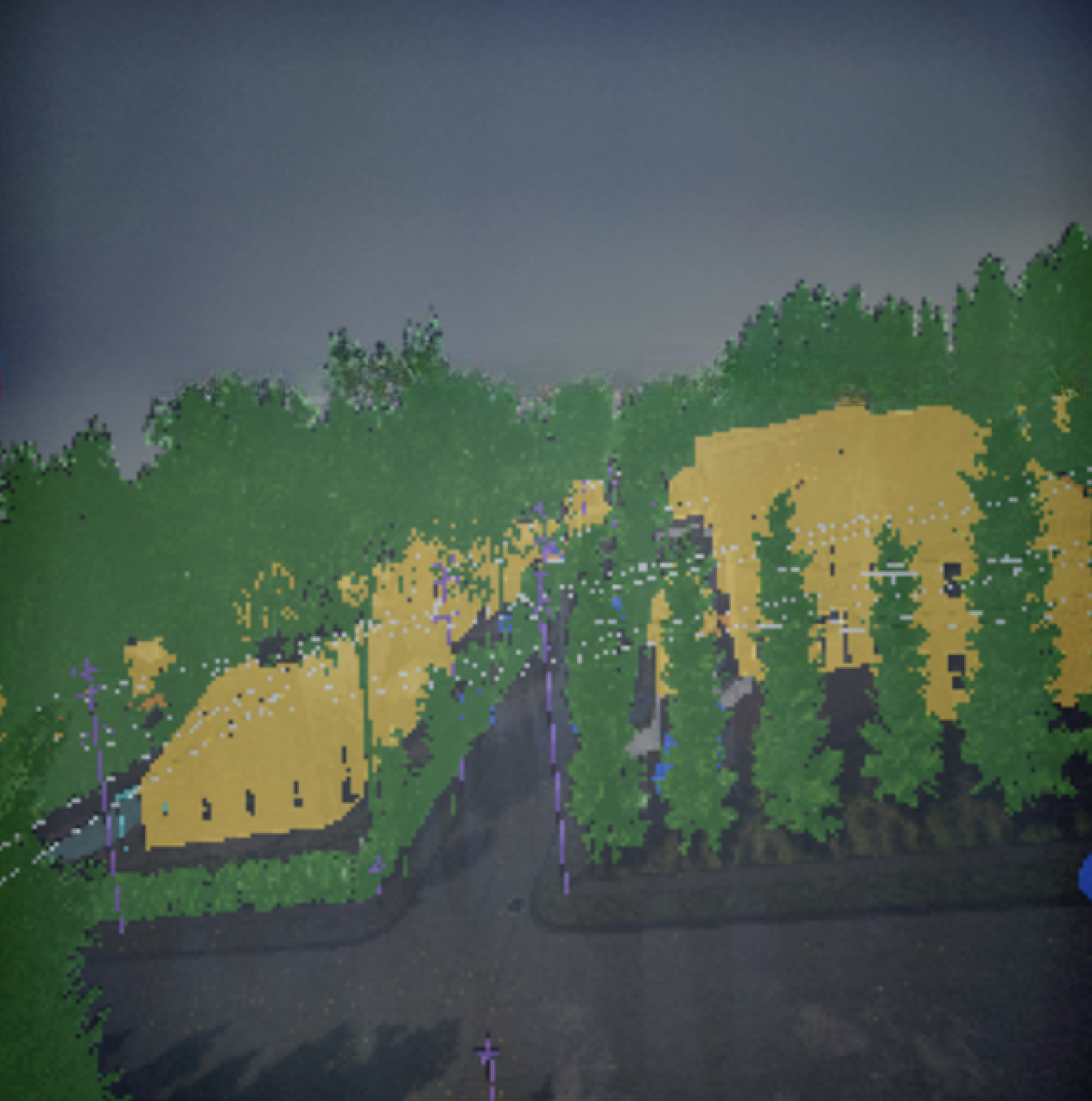}
        \caption{Ground-truth overlay}
        \label{figure:ground-truth}
    \end{subfigure}
    \hfill
    \begin{subfigure}[t]{0.19\textwidth}
        \centering
        \includegraphics[width=\linewidth]
        {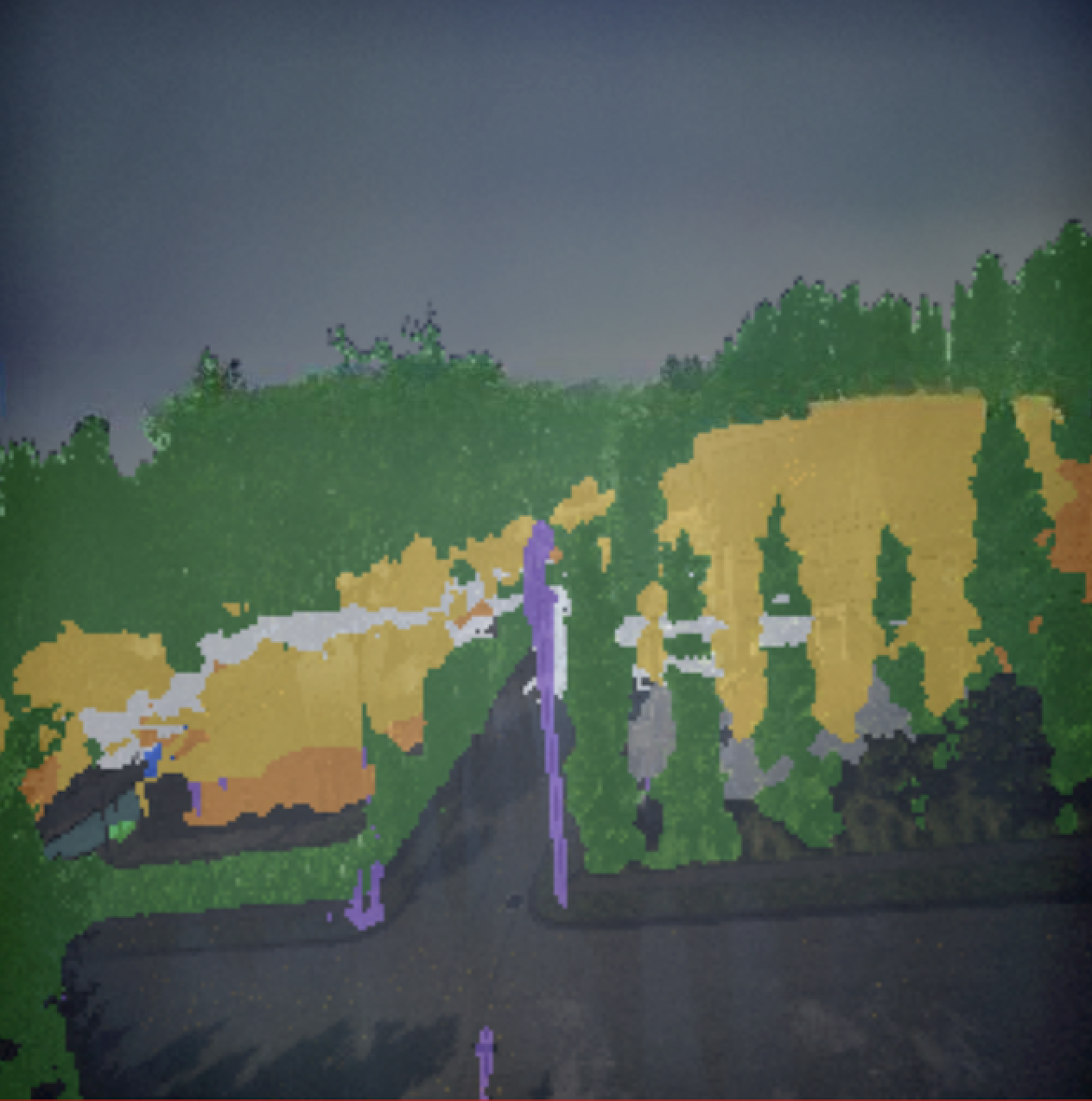}
        \caption{RGB prediction}
        \label{figure:rgb-prediction}
    \end{subfigure}
    \hfill
    \begin{subfigure}[t]{0.19\textwidth}
        \centering
        \includegraphics[width=\linewidth]
        {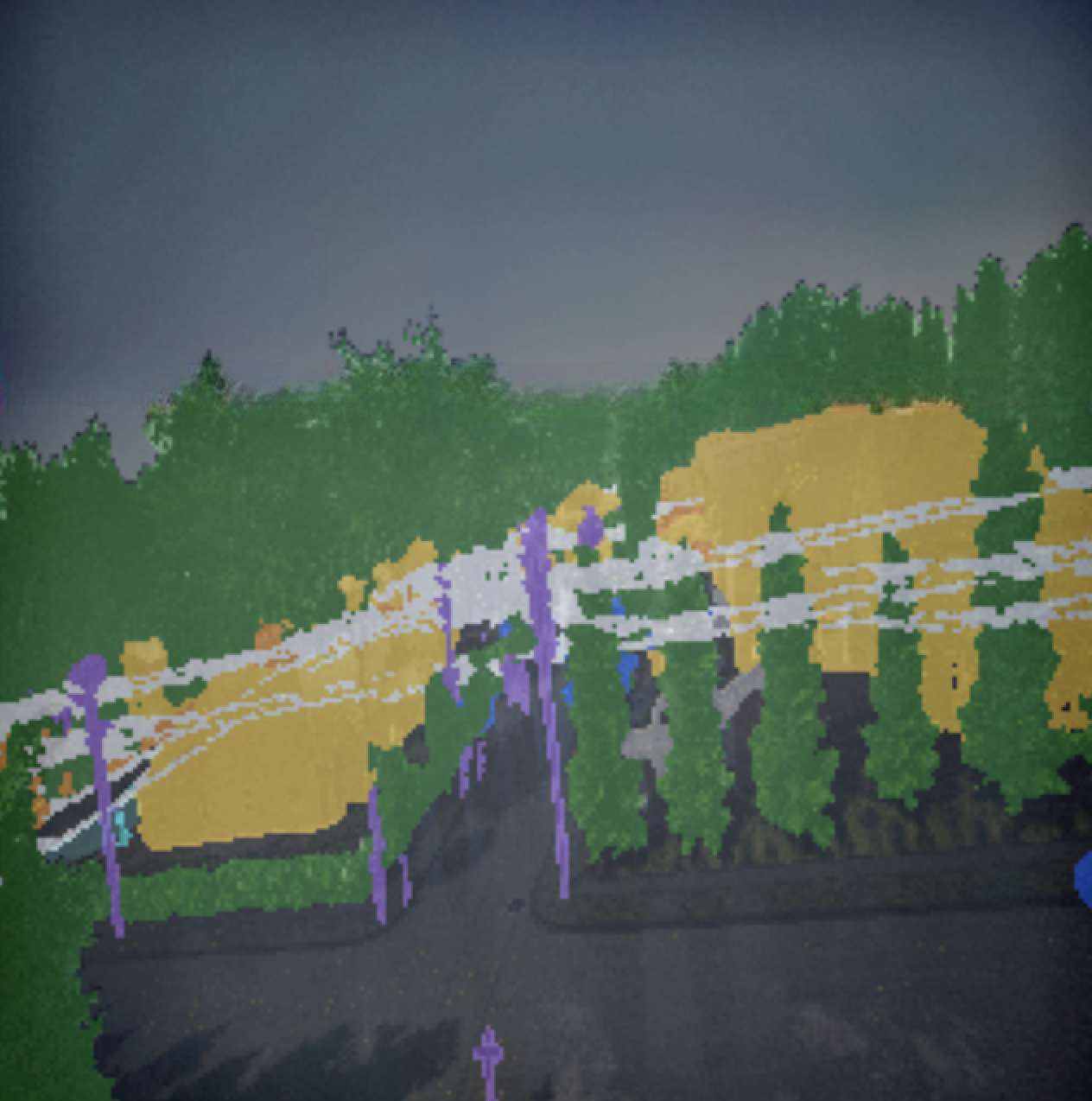}
        \caption{RGB-D prediction}
        \label{figure:rgbd-prediction}
    \end{subfigure}
    \hfill
    \begin{subfigure}[t]{0.19\textwidth}
        \centering
        \includegraphics[width=\linewidth]
        {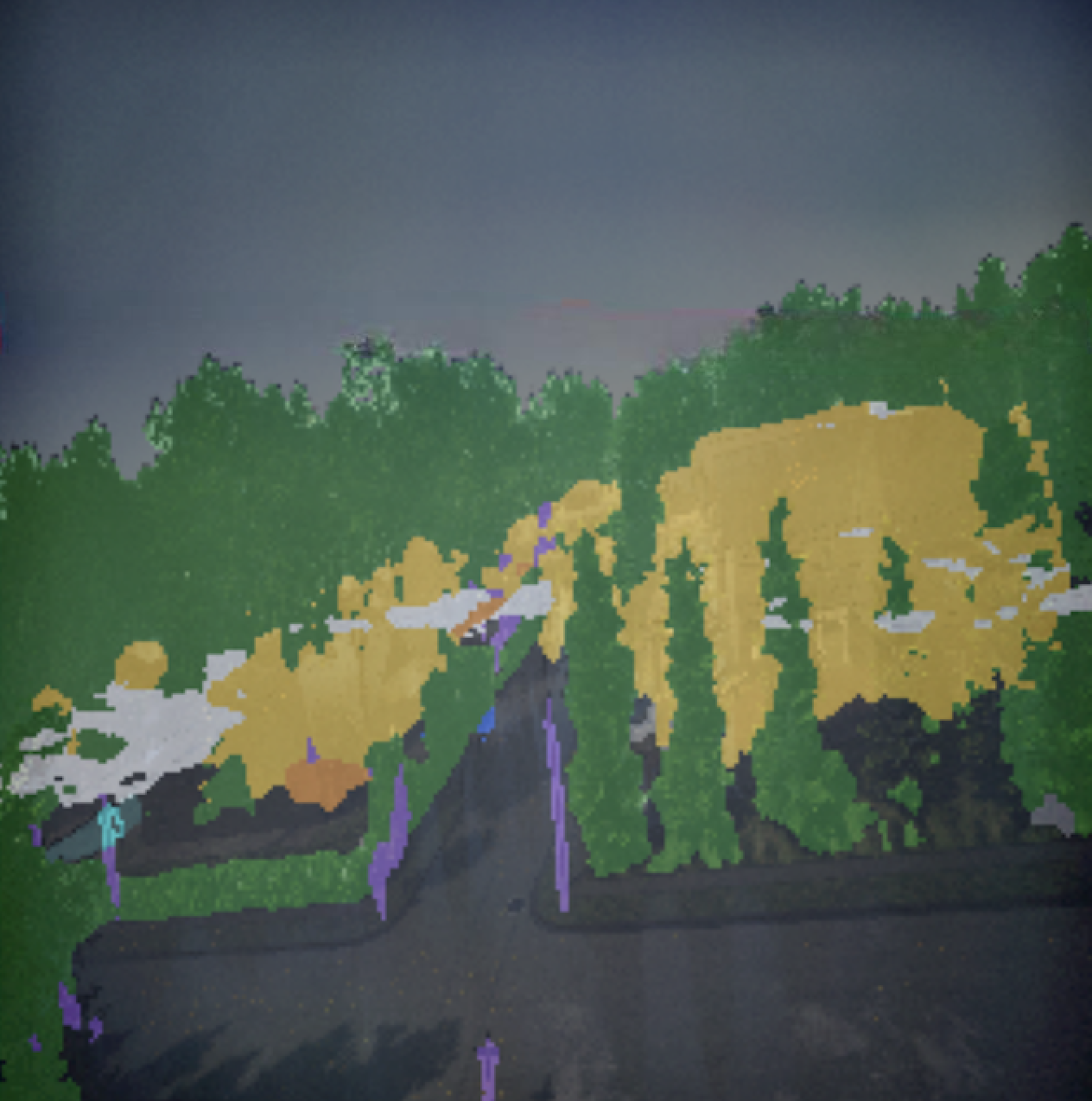}
        \caption{RGB-E prediction}
        \label{figure:rgbe-prediction}
    \end{subfigure}
    \hfill
    \begin{subfigure}[t]{0.19\textwidth}
        \centering
        \includegraphics[width=\linewidth]
        {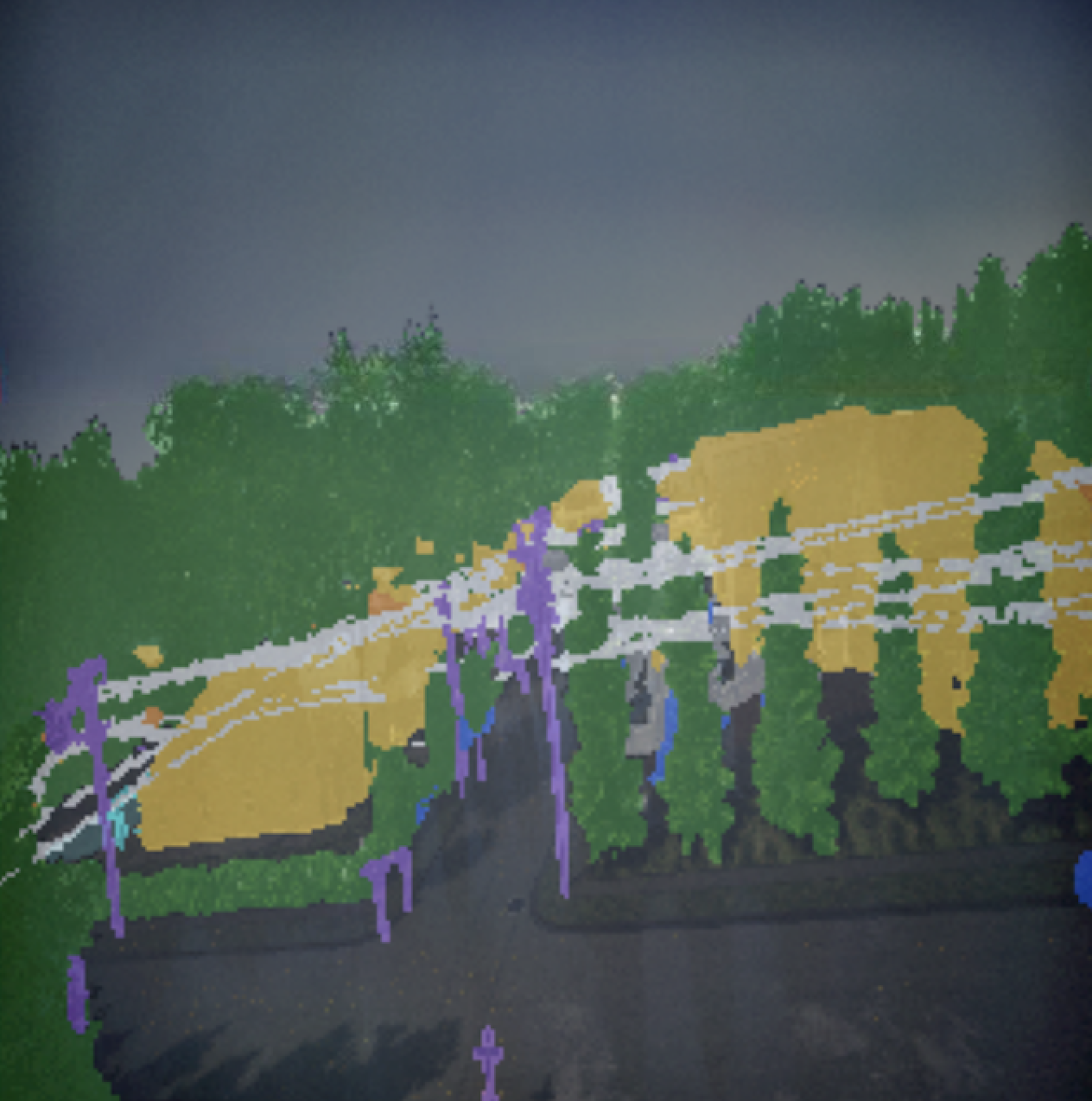}
        \caption{RGB-D-E prediction}
        \label{figure:rgbde-prediction}
    \end{subfigure}

    \vspace{0.4em}


    \begin{minipage}[t]{0.19\textwidth}
        \mbox{}
    \end{minipage}
    \hfill
    \begin{subfigure}[t]{0.19\textwidth}
        \centering
        \includegraphics[width=\linewidth]
        {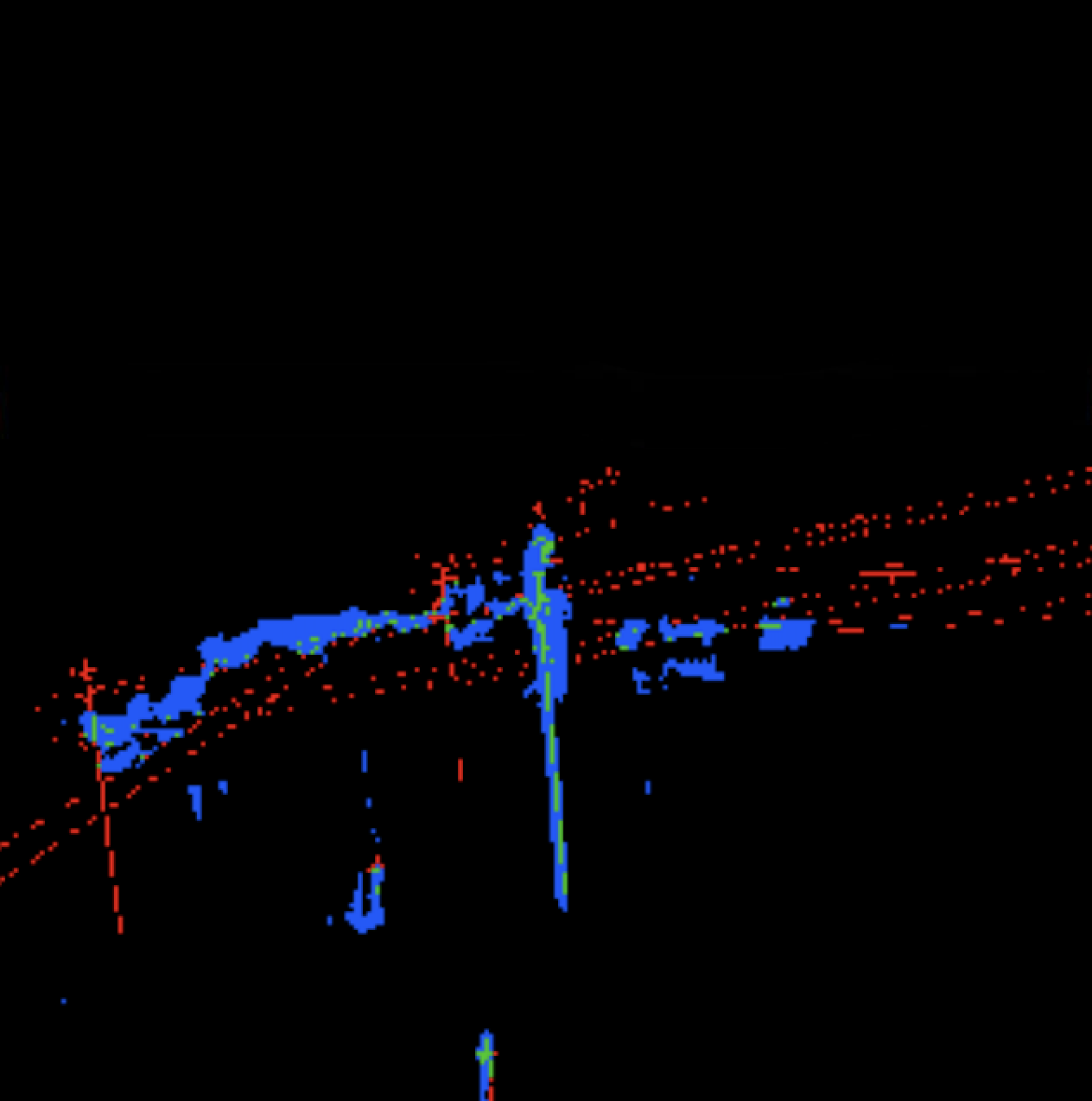}
        \caption{RGB error map}
        \label{figure:rgb-error}
    \end{subfigure}
    \hfill
    \begin{subfigure}[t]{0.19\textwidth}
        \centering
        \includegraphics[width=\linewidth]
        {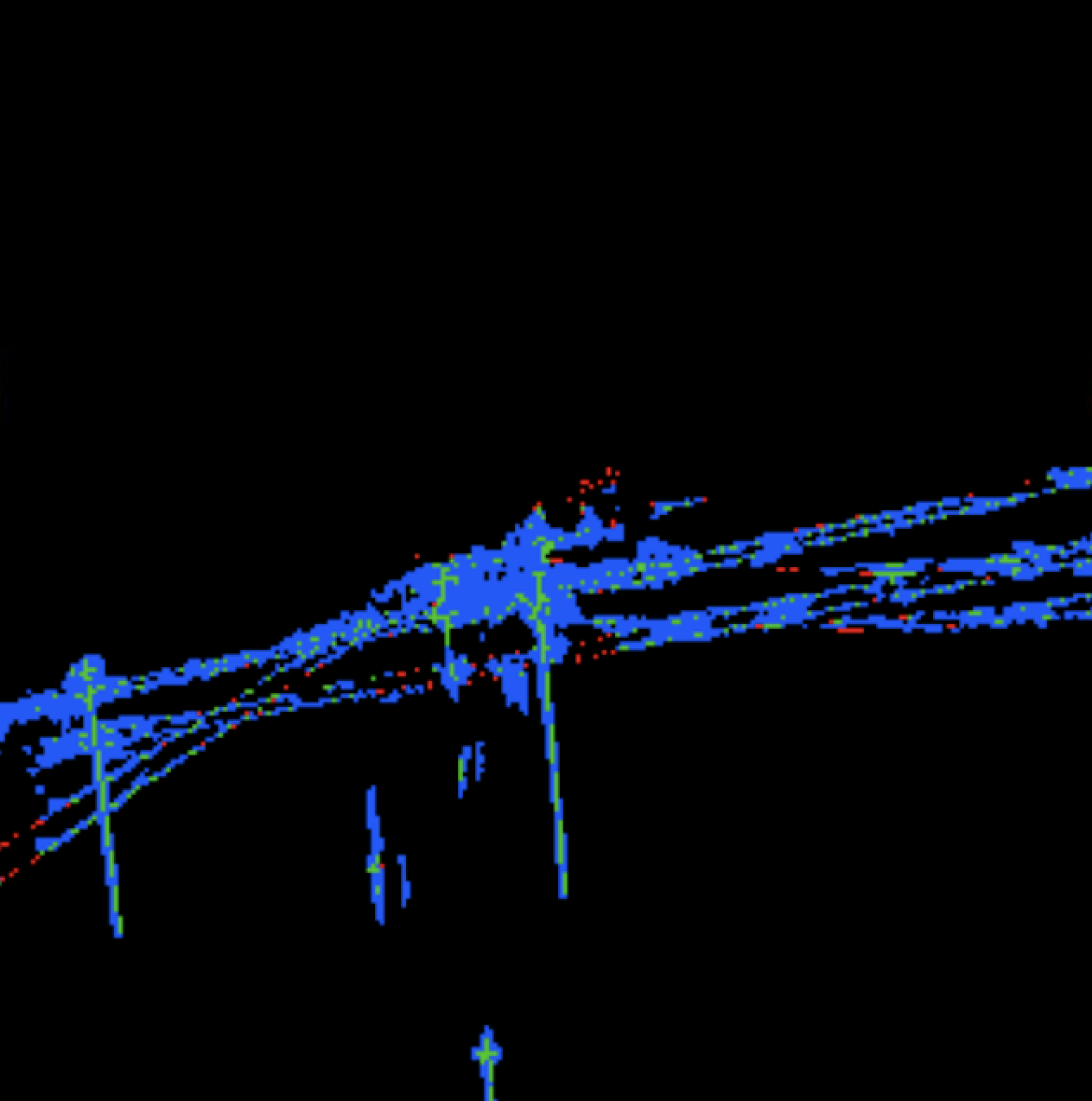}
        \caption{RGB-D error map}
        \label{figure:rgbd-error}
    \end{subfigure}
    \hfill
    \begin{subfigure}[t]{0.19\textwidth}
        \centering
        \includegraphics[width=\linewidth]
        {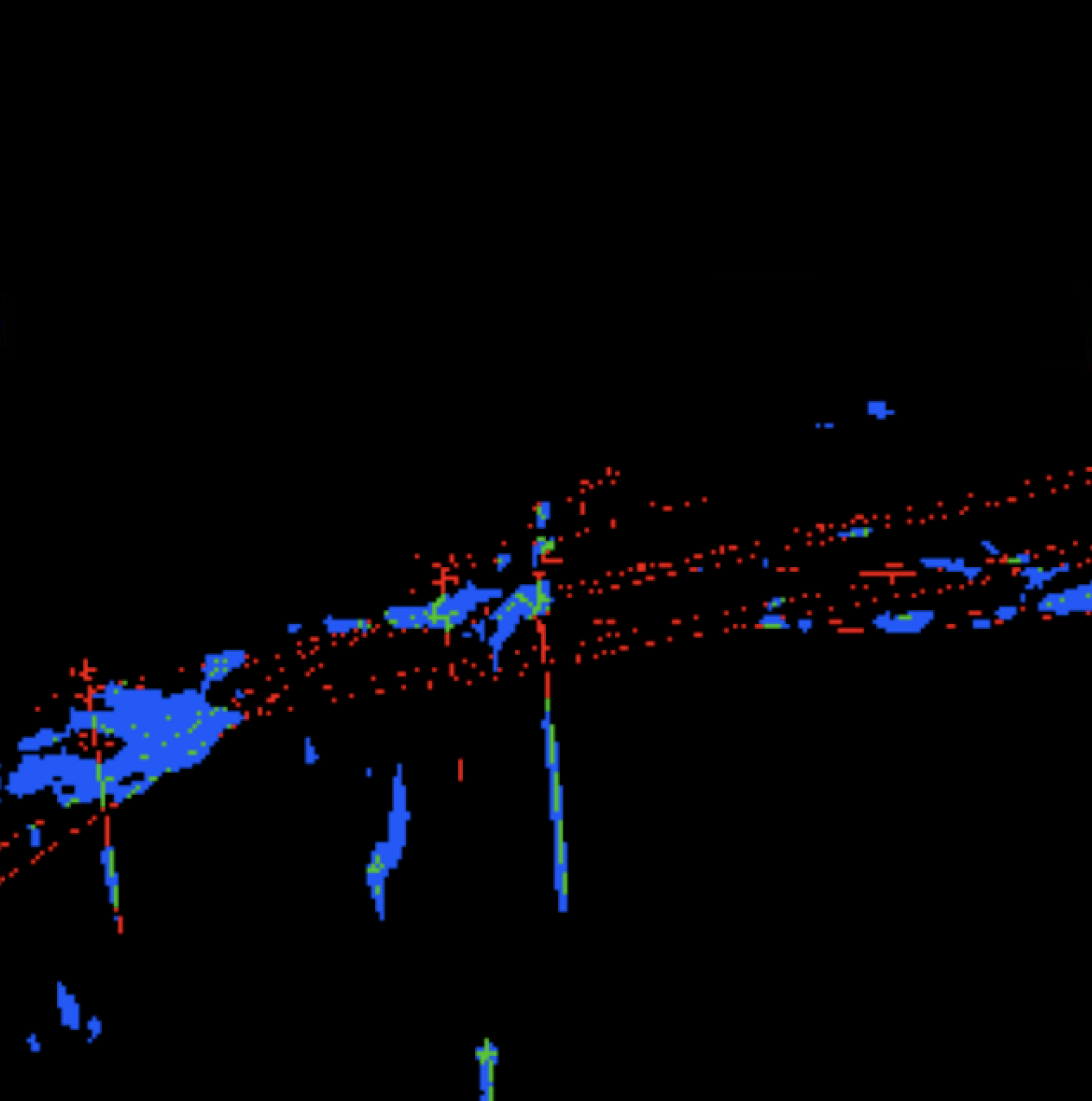}
        \caption{RGB-E error map}
        \label{figure:rgbe-error}
    \end{subfigure}
    \hfill
    \begin{subfigure}[t]{0.19\textwidth}
        \centering
        \includegraphics[width=\linewidth]
        {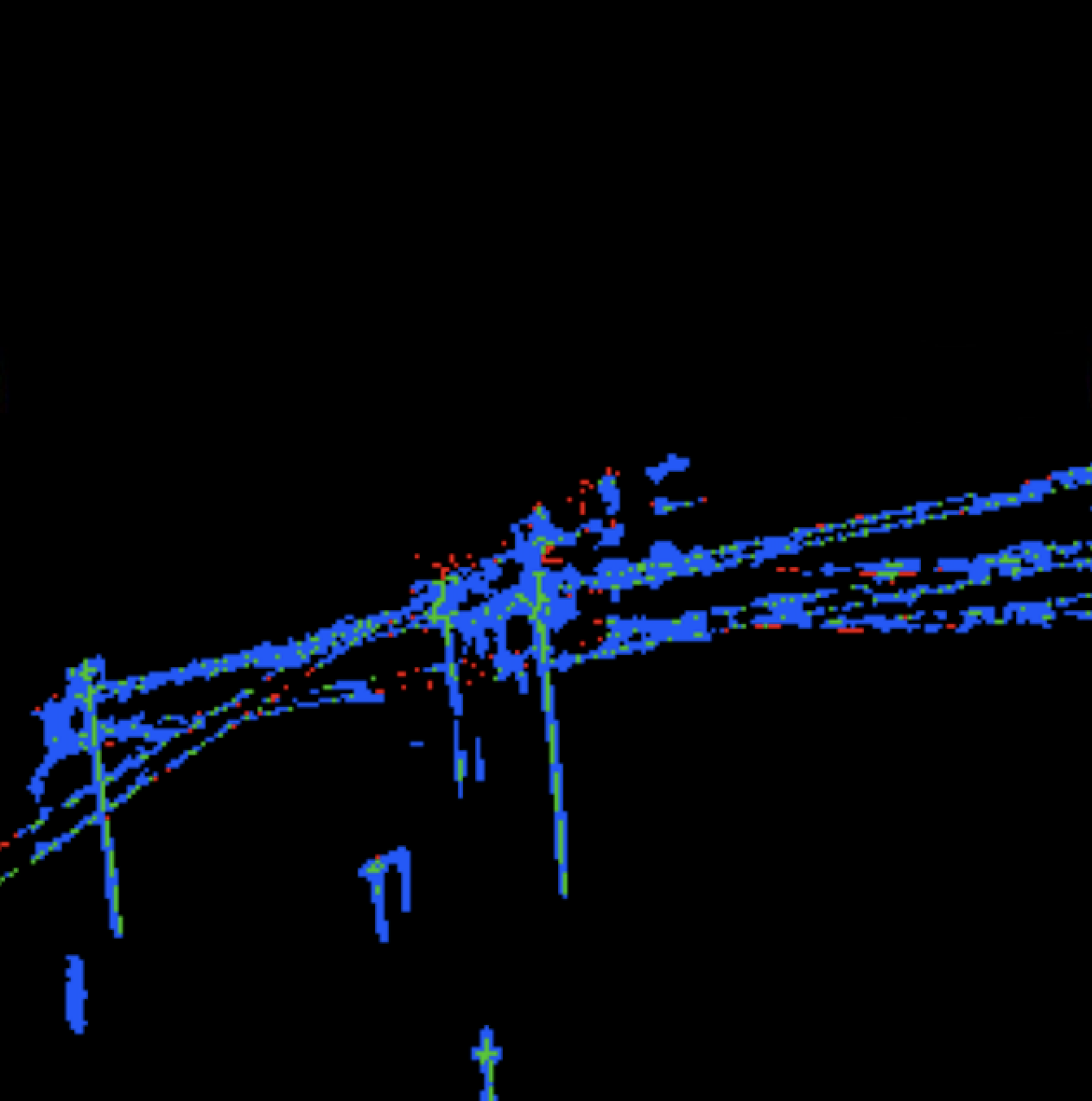}
        \caption{RGB-D-E error map}
        \label{figure:rgbde-error}
    \end{subfigure}

    \caption{
    Controlled qualitative comparison on DDOS test frame \texttt{neighbourhood/12/56} using the non-pretrained U-Net.
    The top row shows the ground truth and prediction overlays across input configurations, while the bottom row shows the corresponding thin-obstacle error maps for the union of \emph{Thin Structures} and \emph{Ultra-thin}.
    Green, red, and blue denote correctly predicted target pixels, false negatives, and false positives, respectively; black denotes all remaining pixels.
    }
    \label{figure:qualitative-results}

\end{figure*}

\noindent\textbf{Input-Cue Effects.}
Depth provides the clearest and most consistent aggregate improvement over RGB. As shown in Table~\ref{table:aggregate-results}(a), RGB-D increases mIoU from 0.4314 to 0.4570, mBIoU from 0.5158 to 0.5781, $\mathrm{IoU}_{\mathrm{Thin}}$ from 0.2068 to 0.2388, $\mathrm{IoU}_{\mathrm{Ultra}}$ from 0.0789 to 0.1185, and mean thin-obstacle $F_2$ from 0.3668 to 0.4602. Table~\ref{table:full-results} further shows that RGB-D exceeds RGB in mIoU, Thin Structures IoU, and Ultra-thin IoU across all eight model settings.

In contrast, RGB-E provides only marginal aggregate changes: mIoU increases from 0.4314 to 0.4317, mBIoU from 0.5158 to 0.5171, and $\mathrm{IoU}_{\mathrm{Thin}}$ from 0.2068 to 0.2072, while Ultra-thin IoU slightly decreases from 0.0789 to 0.0782 and mean thin-obstacle $F_2$ remains nearly unchanged (0.3668 vs.\ 0.3671). Thus, adding edges to RGB yields little systematic aggregate benefit.

Adding edges to RGB-D produces similarly small, metric-dependent changes. RGB-D-E achieves slightly higher mIoU (0.4604 vs.\ 0.4570), mBIoU (0.5807 vs.\ 0.5781), and Ultra-thin IoU (0.1203 vs.\ 0.1185), whereas RGB-D retains higher Thin Structures IoU (0.2388 vs.\ 0.2343) and mean thin-obstacle $F_2$ (0.4602 vs.\ 0.4593). Overall, RGB-D and RGB-D-E are the strongest input configurations, with neither consistently outperforming the other.

\noindent\textbf{Class-wise Cue Effects.}
Across semantic classes, adding depth to RGB is broadly beneficial and consistently improves both Thin Structures and Ultra-thin. As shown in Fig.~\ref{figure:class-wise-results}(a), 76 of 80 model--class $\Delta\mathrm{IoU}$ values for RGB-D relative to RGB are positive. Thin Structures improves in all eight model settings by approximately 0.008--0.051 IoU, while Ultra-thin improves in all eight by approximately 0.029--0.059.

By contrast, RGB-E produces mixed class-wise effects across semantic classes and model settings. Fig.~\ref{figure:class-wise-results}(b) shows both positive and negative $\Delta\mathrm{IoU}$ values, with improvements over RGB in four of eight settings for each of Thin Structures and Ultra-thin, indicating a configuration-dependent edge contribution.

Adding edges to RGB-D remains similarly inconsistent. In Fig.~\ref{figure:class-wise-results}(c), RGB-D-E exceeds RGB-D for Thin Structures in three of eight settings and for Ultra-thin in four of eight, with $\Delta\mathrm{IoU}$ ranges of approximately $-0.024$ to $+0.006$ and $-0.005$ to $+0.019$, respectively. Thus, the incremental contribution of edges remains model- and class-dependent even when depth is available.

\noindent\textbf{Model and Initialization Effects.}
Performance varies across model settings, with the non-pretrained U-Net achieving the highest averages across all five accuracy metrics. As shown in Table~\ref{table:aggregate-results}(b), U-Net attains the highest mIoU (0.5104), mBIoU (0.6091), Thin Structures IoU (0.3197), Ultra-thin IoU (0.1452), and mean thin-obstacle $F_2$ (0.5243). At the configuration level, U-Net+RGB-D achieves the highest mIoU (0.5295), Thin Structures IoU (0.3523), and mean $F_2$ (0.5866), while U-Net+RGB-D-E achieves the highest mBIoU (0.6432) and Ultra-thin IoU (0.1806) (Table~\ref{table:full-results}).

For the architecture-matched DeepLabV3, DeepLabV3+, and SegFormer pairs, pretraining modestly improves all five average accuracy metrics (Table~\ref{table:aggregate-results}(b)). mIoU increases from 0.4286 to 0.4352, 0.4286 to 0.4412, and 0.4245 to 0.4346, respectively, while mean thin-obstacle $F_2$ increases from 0.3786 to 0.3907, 0.3874 to 0.4071, and 0.3742 to 0.3951. These gains are modest and architecture-dependent. U-Net is excluded from this initialization-only comparison because its implementations are not architecture-matched.

\noindent\textbf{Efficiency--Accuracy Trade-offs.}
Inference efficiency varies substantially more across model settings than across input configurations. In Table~\ref{table:aggregate-results}, average FPS spans 647.4--690.6 across inputs in Subtable~(a), versus 129.6--1419.5 across model settings in Subtable~(b), with no systematic reduction in average throughput from adding depth or edge cues. The non-pretrained U-Net achieves the highest average accuracy across the reported metrics but the lowest throughput (129.6 FPS), illustrating a model-level accuracy--efficiency trade-off.

\noindent\textbf{Qualitative Analysis.}
For the DDOS test frame \texttt{neighbourhood/12/56} in Fig.~\ref{figure:qualitative-results}, the non-pretrained U-Net recovers thin obstacles more completely with RGB-D and RGB-D-E than with RGB and RGB-E, although false-positive and false-negative regions remain. This visual pattern is consistent with Table~\ref{table:full-results}, where mean thin-obstacle recall/$F_2$ is 0.6538/0.4705 for RGB, 0.8555/0.5866 for RGB-D, 0.6332/0.4595 for RGB-E, and 0.8474/0.5807 for RGB-D-E. The example therefore illustrates improved target recovery with depth while residual segmentation errors remain.

\section{Discussion}
\label{section:discussion}
The results identify depth as the most consistently useful auxiliary cue in the evaluated early-fusion setting, improving overall segmentation and both thin-obstacle classes across model settings, whereas Sobel edges produce only marginal and configuration-dependent effects. The mixed performance of RGB-D and RGB-D-E further indicates that cue benefits are not uniformly additive and depend on the metric, semantic class, and model setting. Architecture also has a substantial influence on performance: the non-pretrained U-Net achieves the highest averages across the reported accuracy metrics, while pretraining yields modest gains for the architecture-matched DeepLabV3, DeepLabV3+, and SegFormer pairs. Throughput varies substantially more across model settings than across input configurations, illustrating a model-level accuracy--efficiency trade-off. The qualitative example is consistent with these trends, showing improved thin-obstacle recovery for depth-enabled inputs while false-positive and false-negative regions remain. Overall, the findings support depth-enhanced early fusion as a simple baseline for UAV thin-obstacle segmentation, while showing that performance remains sensitive to model choice, obstacle category, and cue configuration.

\section{Conclusion and Future Work}
\label{section:conclusion}
This work presented \tool, a modular early-fusion framework for studying the individual and combined effects of depth and explicit edge information relative to RGB in UAV thin-obstacle semantic segmentation. Evaluation on DDOS across multiple input configurations, architectures, and initialization settings shows that the benefits of auxiliary inputs are not simply additive and vary across model settings and obstacle categories. By keeping the fusion mechanism fixed, \tool provides a controlled baseline for assessing cue utility independently of fusion complexity and can inform the design and evaluation of more specialized multi-cue fusion strategies.

\label{section:future-work}
Future work will extend this controlled analysis to additional datasets and repeated runs to assess the robustness of the observed patterns. Building on the fixed early-fusion baseline, more adaptive fusion strategies, including attention-based, gated, intermediate, and late fusion, can then be systematically evaluated to determine whether complementary information can be exploited more effectively. The class-dependent behavior observed for \emph{Thin Structures} and \emph{Ultra-thin} obstacles also motivates class-aware objectives and representations. Finally, evaluation on UAV deployment hardware will help determine how the observed segmentation and efficiency characteristics translate to practical operating conditions.

\section*{Data Availability}
\label{section:data-availability}
The Drone Depth and Obstacle Segmentation (DDOS) dataset used in this study is publicly available through its official release~\cite{kolbeinsson2024ddos}. The EDFNet source code and reproducibility materials are publicly available on GitHub at \url{https://github.com/negarfathi/EDFNet} and archived on Zenodo~\cite{edfnet-artifact}.

\section*{Acknowledgments}
\label{section:acknowledgments}
The initial motivation for this research arose from a robotics course project. The author thanks Brittany Duncan for her guidance and feedback, and Roxana Shajarian and Mohammad Jalili Torkamani for their contributions to that earlier course project.

The author acknowledges the use of OpenAI's ChatGPT (GPT-4o)~\cite{chatgpt} solely for grammar checking, language editing, and clarity improvement. All incorporated edits were reviewed and approved by the author, who takes full responsibility for the final manuscript.

{
    \small
    \bibliographystyle{ieeenat_fullname}
    \bibliography{references}
}

\end{document}